# 3D TERRAIN SEGMENTATION IN THE SWIR SPECTRUM

*Dalton Rosario[1], Anthony Ortiz[2], and Olac Fuentes[2]*

U.S. Army Research Laboratory[1], University of Texas at El Paso[2]

## ABSTRACT

We focus on the automatic 3D terrain segmentation problem using hyperspectral shortwave IR (HS-SWIR) imagery and 3D Digital Elevation Models (DEM). The datasets were independently collected, and metadata for the HS-SWIR dataset are unavailable. We explore an overall slope of the SWIR spectrum that correlates with the presence of moisture in soil to propose a band ratio test to be used as a proxy for soil moisture content to distinguish two broad classes of objects: live vegetation from impermeable manmade surface. We show that image based localization techniques combined with the Optimal Randomized RANdom Sample Consensus (RANSAC) algorithm achieve precise spatial matches between HS-SWIR data of a portion of downtown Los Angeles (LA (USA)) and the Visible image of a geo-registered 3D DEM, covering a wider-area of LA. Our spectral-elevation rule based approach yields an overall accuracy of 97.7%, segmenting the object classes into buildings, houses, trees, grass, and roads/parking lots.

## 1. INTRODUCTION

There are many challenges associated with the fusion of different sensing modalities, to include the uncertainty of feature correspondence among the modalities due to differences in measured phenomena, spatial resolutions, and viewing perspectives [1]-[2]. We focus on the problem of enabling a machine to automatically segment a digitized terrain by first fusing aerial Shortwave infrared (SWIR: 1.0-2.5 μm) hyperspectral data with 3D Lidar-derived Digital Elevation Models (DEMs) that overlap the target scene, then employing a spectral-elevation rule based approach to segment the digitized terrain into finer material classes (e.g., high buildings, houses, tree clusters, grassy areas, roads/parking lots). Reliable performance of such automatic tasks through software is considered useful for various types of commercial and military applications, to include mission planning for rescue operations [3]. We further constrain the problem in twofold: (i) by using unregistered datasets from independent data acquisitions and (ii) by not relying on the geo-registration of data from one of the sensing modalities. The problem addressed in this paper departs from prior works on 3D terrain segmentation, which exclusively use hyperspectral data in the visible spectrum (Vis: 0.4-0.7 μm) that are already spatially registered to 3D DEMs (see past IGARSS challenges [4]), where variations to the state of the art machine learning methods are applied to segment the digitized scene via object classification. This paper also explores an overall slope observed in SWIR spectra that correlates with the presence of moisture in soil to propose a band ratio test to be used as a proxy for soil moisture content that can distinguish two broad classes of material types: live healthy vegetation from impermeable manmade surface.

## 2. SWIR SPECTRAL ANALYSIS FOR MATERIAL SEGMENTATION

The products derived from both spectral and spatial analyses that are helpful to mission planning consist of three procedures: Determination of soil moisture content; segmentation of the image into vegetation, soil, and water; and derivation of lines of communication (roads, rivers). The amount of moisture in soil can impact a number of phenomena important for mission planning, including the amount of dust in the air and the firmness of the ground. It is intuitive that should the soil exceed a certain level of moisture content, ground mobility of tracked and wheeled vehicles will be adversely affected. Remotely sensed spectral imagery can be used to determine the relative amounts of moisture in surface soils by examining the reflectance in the SWIR region of the spectrum. The atmosphere is opaque to SWIR radiation near 1.4 μm and 1.9 μm due to the presence of gaseous water and carbon dioxide. As a result, surface properties at these wavelengths are not accessible from remotely sensed images. However, immediately adjacent to these regions, the spectrum is influenced by the presence of absorbed water (see Fig 1). This is true for an absorption feature near 2.2 μm as well. As a result of the analysis of laboratory spectra of three types of soil—characterized as loam, clay, and sand—it was observed that the overall slope of the spectrum in the SWIR also correlates with the presence of moisture in the soil. Specifically, in terms of the hyperspectral SWIR bands used in for this paper, dry soils exhibited an increase in reflectance in bands between 2.09 μm and 2.15 μm over bands between 1.55 μm and 1.75 μm, see Fig. 1. Wet soils were either flat or decreased in reflectance over that span.

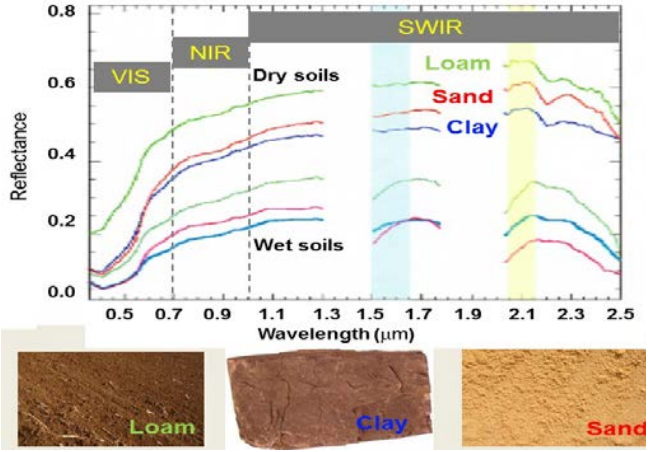

Fig. 1. Laboratory reflectance spectra for three types of soils—loam (green), sand (red), and clay (blue)—under two moisture conditions. Areas most sensitive to the presence of liquid water are highlighted by light blue and light yellow vertical bands. Note the contrast between dry and wet samples. Material spectra courtesy of Spectral Information Technology Application Center Library.

This suggests that a band sub-range ratio test can be used as a proxy for soil moisture content in the HS SWIR image used in our work. We explore this phenomenon by proposing the following wetness index in order to distinguish two broad classes of objects in a given terrain (live healthy vegetation from impermeable manmade surface):

$$Ratio_{WET} = \frac{\sum_{j=1.55\mu m}^{1.75\mu m} d_j}{\sum_{j=2.09\mu m}^{2.35\mu m} d_j}, \qquad (1)$$

where $d_j$ is the radiance or reflectance value at band $j$. Notice from Fig. 1 that the inverse of (1), i.e., 1/*Ratio*, is expected to yield real values between 0 and unity ([0,1]) for wet materials. The behavior of such a random variable may be modeled by a flexible distribution defined on the interval [0,1], such as the beta distribution function, see Fig. 2.

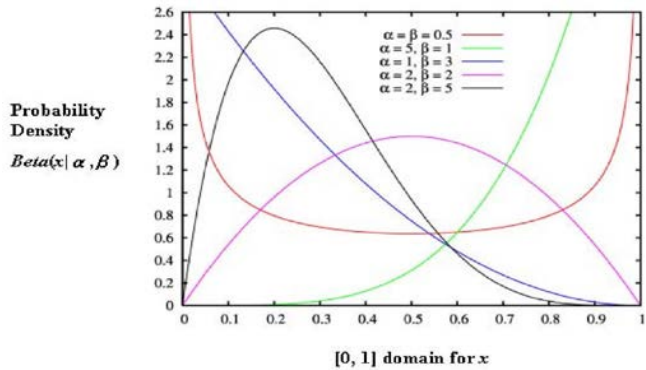

Fig. 2. Beta probability density function.

The beta distribution is a family of continuous probability distributions defined on the interval [0, 1] parametrized by two positive shape parameters, denoted by $\alpha$ and $\beta$, that appear as exponents of the random variable and control the shape of the distribution. Modeling (1) as a random variable and restricting its behavior by the beta distribution family to determine the material wetness concentration would allow a likelihood ratio test to be formulated. It is beyond the scope of this paper to formulate this hypothesis test, although we do have interest in pursuing this approach in the future. In the meantime, we anticipate that exploring the wetness index in (1) with an adaptively estimated threshold can broadly segment the image between live vegetation and manmade structures in imagery, as it will be shown later. Object class finer distinction (e.g., between trees and grass) can further be segmented by aligning elevation data to the segmented map pixels. Results using (1) and elevation measurements are also discussed later.

## 3. DATASETS

We used a dataset collected by Headwall Photonics over a multi-block urban area of downtown Los Angeles (LA), California, USA, using Headwall's Hyperspec® SWIR hyperspectral imaging sensor [5] onboard a small manned airplane. Key sensor specifications: 384 spatial bands, 260 spectral bands, wavelength range 0.9-2.5 μm, maximum frame rate 450 Hz, Stirling-cooled MCT detectors. The manned airplane flew the *pushbroom* hyperspectral imaging system over the target area in Los Angeles, collecting 829 samples by 260 SWIR bands (defined in this case as a frame) and using the airplane's movement to obtain 1,163 lines. A representation of the datacube consisting of 1,163 lines by 829 samples by 260 bands of the target area in Los Angeles is shown as the band average (less than 1 m pixel resolution) in Fig. 3. The area represented in Fig. 3 includes live vegetation (trees, grass) and manmade structures (high buildings, houses, roads, and parking lots, and other unknown manmade objects). Notice in Fig. 3 that the data acquisition occurred closer to sunset; explaining the prolonged shadows observed near the high buildings. We conjecture that scale invariant features of terrain landmarks between SWIR images and spatially corresponding Vis images will be highly correlated; thus, we decided to explicitly explore this intuition as part of our image registration approach. For elevation measurements (Fig. 4), we used a 3D DEM representing a significantly wider area of Southern California, containing a large portion of LA County, to include mountains and the entire downtown LA.

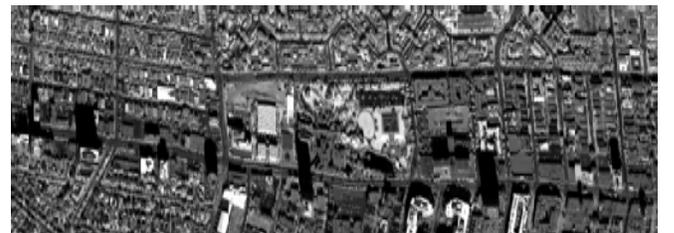

Fig. 3. SWIR hyperspectral band average of a multi-block urban region of Los Angeles, California, USA.

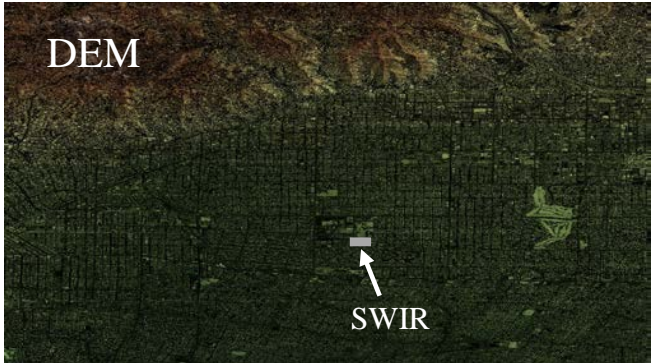

Fig. 4. 3D DEM (1 m pixel resolution) of a portion of Los Angeles County, California, USA, which includes in relative scale the multi-block downtown portion of Los Angeles where the SWIR hyperspectral data were collected.

The DEM is geo-registered for Latitude and Longitude world coordinates per pixel of equal size (1 m spatial resolution), and includes both elevation measurements from a Lidar system and a corresponding geo-registered orthophoto (nadir Vis image) to put scene context in the DEM. Fig. 4 depicts the DEM and shows in relative scale the spatial area where within the DEM spatial area the SWIR data acquisition occurred; but of course this spatial location was unknown prior to the execution of our image registration method using these two independent regions of the electromagnetic spectrum: SWIR and Vis. The provided orthophoto with the DEM also covers an area that is many orders of magnitude larger than the area covered by the SWIR datacube. In the scope of this paper, successful registration by a machine using these datasets constitutes a successful automatic association of elevation data with spectra, given that for each aligned pixel between these two images there is a corresponding spectrum available in the hyperspectral datacube. Our image registration approach is discussed next.

### 4. SWIR-VISIBLE SPATIAL MATCHING

Inspired by image-based localization approaches similar to the one introduced by Sattler et al. in [6], and the Optimal Randomized RANSAC algorithm [7], we succeeded in registering the 2D spatial area of the SWIR hyperspectral datacube onto the 3D DEM using a variation of our prior work [8], resulting in the association of elevation measurements with spectra; as follows. The fusion approach starts by representing each 3D point in the DEM by all the Scale Invariant Feature Transform (SIFT) descriptors [9] contributing to its spatial locations in the Vis image. All the descriptors of a single 3D point are assigned to a visual word. The K-means clustering algorithm is applied to cluster all the 3D point clouds into k clusters. We used 100,000 clusters in this experiment. To improve computational efficiency, we avoid comparing all of the points in the point clouds by assigning the centroids obtained from the k-means to be the visual words. SIFT is independently computed on the SWIR and Vis reference images and results are represented as visual words for comparison with visual words from the previous step. The visual word comparison strategy can then be reduced by finding through a sequential search the two nearest visual words to the visual word (hyperspectral-based SIFT descriptor). The sequential search yields a set of correspondences between each hyperspectral-based SIFT descriptor and two 3D model visual words. A correspondence is accepted if the two nearest neighbors pass the SIFT ratio test using a threshold. This is based on the idea that the probability that a match is correct can be determined by taking the ratio of distance from the closest neighbor to the distance of the second closest [9]. The method rejects all matches in which the distance ratio is greater than 0.7. This threshold works empirically well for this dataset and many others. If more than one 2D feature matches exist in association with the same 3D point, the descriptor with smallest Euclidean distance is selected. The linear search continues until a user-specified number of correspondences is satisfied (in this experiment this number was set to 100), or the search is exhausted from the hyperspectral data perspective. The resulting set of correspondences does not necessarily guarantee a geometric alignment of the hyperspectral images onto the 3D DEM that would match the quality that could have been achieved by human intervention. In order to improve the alignment process, we applied the Optimal Randomized RANSAC algorithm to the resulting set of correspondence vectors obtained up to this stage. The geometric alignment is declared as acceptable using the criterion that more than $n$ correspondences ($n = 5$ works well empirically in this dataset) must be inliers to accept a match.

### 5. RESULTS AND DISCUSSION

The fusion approach described in Section 4 was applied to the hyperspectral SWIR dataset and Vis orthophoto of corresponding DEM using the following details. We chose band 14 (near 1.2 μm) to represent the spatial area of the entire SWIR hyperspectral datacube because of the high reflectance property of most material types in an urban scene, relative to other bands. In order to address more efficiently the computational load and associated time, given the vast spatial coverage of the DEM relative to the smaller hyperspectral SWIR spatial coverage (see Fig. 4), we split in half the spatial area of the SWIR band-14 and allowed each half to compute their spatial matches independently of each other, such that each half estimated its spatial feature matches within a neighborhood centered at a random (without replacement) spatial location within the wide-area Vis orthophoto, thus eliminating once for all the undesirable locations in the wide-area digitized scene and retaining the better prospects. This strategy is very effective because it allows the computational load, which is in the order of $O^2$ relative to the spatial area, to be processed in parallel using two smaller areas rather than simultaneously using a single twice-as-large spatial mosaic, cutting in half the matching-search computational time. This matching search strategy paid off and yielded the result depicted in Fig. 5, where the segmented (white, yellow, red, dark) map represents the SWIR spatial area overlaid onto the larger local neighborhood in the Vis orthophoto that produced the best matching result. This segmented map was the result from applying the wetness spectral ratio test in (1), to be discussed shortly. Notice in Fig. 5 (segmented map) that the split match is slightly misaligned, since they were performed independently from each other. We could rectify this misalignment by applying a final matching search as an additional last step between the entire SWIR mosaic and the neighborhood where this split conversion occurred in the Vis orthophoto; this optimization, however, was not done for this paper. The real advantage from this spatial match is that each voxel in the SWIR datacube not only features 270 bands but also has a corresponding elevation value from the DEM. Using this data association, it will be possible to refine the terrain segmented maps based on spectral and elevation properties as discussed earlier.

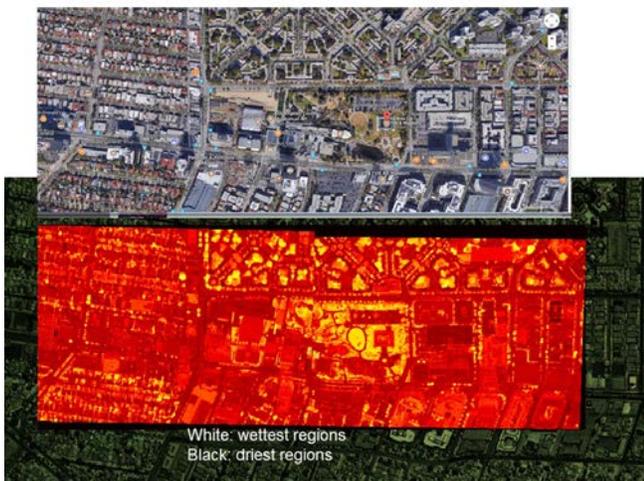

Fig. 5. Material segmented map separating material types in the scene featuring large concentration of water (trees, bushes, and grass), they are represented by white-yellow pixels in the center image. Red to darker colored pixels indicate the presence of impermeable manmade surfaces in the scene (buildings, houses, roads, and other unknown manmade objects. Top image is a corresponding Google Map RGB image.

The top image in Fig. 5 is the corresponding *Google Map* Vis photo of the target LA area, which was taken in July 2017—this is about 24 months after the HS SWIR data acquisition of the same area occurred, and is being used in here for visual reference for the reader to qualitatively check the material types present in the scene. Immediately below the Google Map Vis photo is the overlaid segmented map, indicating the materials in the scene containing large concentrations of water, where white and yellow pixels depict the wettest materials and red and darker pixels depict the driest materials. A visual inspection of Fig. 5 suggests that the material types in the scene featuring large concentration of water are those materials composed of live vegetation (i.e., trees, bushes, and grass), which are represented by white-yellow pixels in the segmented map. Pixels showing red to darker colors seem to belong to manmade surfaces in the scene, i.e., buildings, houses, roads, parking lots, etc. Notice also from the Google Map Vis photo that the level of wetness observed in the segmented map seems agnostic to the actual color of the objects consisting of live vegetation in the scene, e.g., both brownish grassy regions and green tree clusters observed in the Google Map Vis photo are indistinguishable in the segmentation map. This result suggests that the wetness spectral ratio in (1) is an effective means to separate live vegetation from manmade structure in the scene, a crucial capability for obtaining mission planning geospatial products. Fig. 6 includes results with elevation.

Fig. 6 depicts the result from our spectral-elevation rule based segmentation method, where by merely using the spectral property distinction in (1) between live vegetation and manmade structure in the scene; and taking into account the voxels' corresponding elevation data, which was possible because of the successful registration between the HS SWIR datacube and DEM, a visual inspection between the segmented map in Fig. 6 and the Google Map color image in Fig. 5 suggests that *red* objects in the segmented map represent buildings (yielded low wetness index) greater than a typical two floor house (ground level included) in the United State of America (USA), *white* objects represent neighborhood houses (also having low wetness index) of lower height than three floors, *green* objects are the grass regions and other lower canopy (e.g., bushes), *yellow* objects represent the trees (having both high wetness index and an elevation greater than shorter canopy), and finally *black* regions represent the roads (having low wetness index and low elevation), parking lots, and other unknown manmade objects having a lower elevation than a typical single-floor house in the USA (e.g., motor vehicles). It is worth noting that the results depicted in Fig. 6 were obtained without the shortfalls associated with supervised machine learning, where a vastly large database of labeled samples is required and often unavailable, and with statistical clustering, where prior information about the total number of individual clusters are often required. In our approach, the wetness index in (1) was computed for the entire digitized scene, a wetness threshold was applied (adaptively estimated to be halfway between the lowest and highest values from the scene), and finally using the corresponding elevation data and two elevation thresholds (lower than a single floor house and higher than a two-floor house), the resulting terrain segmented map is obtained as an useful mission planning product.

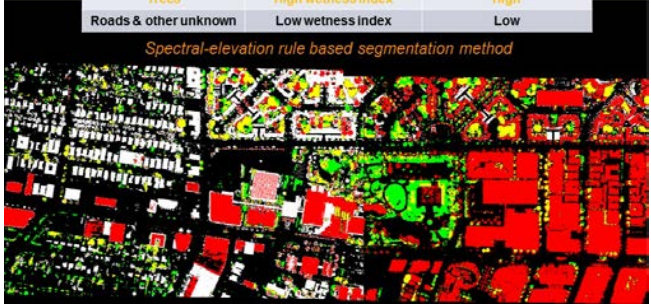

Fig. 6. Segmented map of key objects in the scene produced by our spectral-elevation rule based method, highlighting tall buildings (red), houses (white), trees (yellow), and roads/parking lots/other unknown manmade objects (black).

To quantify the approach's performance, we withdrew from the SWIR datacube and stored a sample library of key material classes (i.e., tree clusters, grassy regions, buildings, houses, roads/parking lots), such that each class consisted of 500 randomly selected spectral samples, without replacement, with associated elevation datum per spectral sample. We applied the wetness index in (1) and the spectral and elevation thresholds used to generate the segmented map depicted in Fig. 6 in order to produce the performance results shown in Table 1. Table 1 summarizes performance in terms of *Precision*, *Recall*, and *Accuracy*, which are borrowed from mathematical statistics and have become popular in the computer vision, document retrieval, and medical test communities [6]. *Precision* only takes into account the information contained within the predicted class columns, yielding a normalized correct detection in terms of false alarms per predicted class that is not constrained by the true sample size per true class. *Recall* takes only into account the information contained within the *true class* row, also yielding a normalized correct detection in terms of missing the target class; notice that each row sum is constrained by the sample size of each true class, in this case

**Table 1. Quantified Performance.**

| Predicted Class / True Class | Tree | Grass | Building | House | Road/ Parking Lot |
|---|---|---|---|---|---|
| Tree | 478 | 1 | 8 | 13 | 0 |
| Grass | 3 | 491 | 0 | 0 | 6 |
| Building | 54 | 0 | 446 | 0 | 0 |
| House | 7 | 4 | 41 | 448 | 0 |
| Road/ Parking Lot | 0 | 5 | 0 | 0 | 495 |
| Precision (%) | 88.1 | 98.0 | 90.1 | 97.2 | 98.8 |
| Recall (%) | 95.6 | 98.2 | 89.2 | 89.6 | 99.0 |
| Accuracy (%) | 96.6 | 99.2 | 95.9 | 97.4 | 99.6 |

500 samples per class. Finally, *Accuracy* does give a truly comprehensive performance measure taking into account all of the key counts in a confusion matrix (i.e., a measure of everything that went right normalized by everything that went both right and wrong in the test trial). The spectral-elevation rule based approach did perform well for this set of metrics, and very well for the overall accuracy at 97.7%. This overall performance measure was computed by taking into account all of true/false positives and true/false negatives from all classes. Since elevation data were used to distinguish the classes, in addition to the wetness index in (1), we would not expect to observe false alarms between classes that fall outside expected elevation ranges (e.g., *Building* and *Road/Parking Lot*), although the *Building* false alarm count of 41 due to *House* samples (see Table 1) suggests that in some of the buildings' portions in the scene fall within the elevation range of the house class.